\documentclass[11pt,a4paper]{article}
\usepackage[hyperref]{acl2021}
\usepackage{times}
\usepackage{latexsym}

\usepackage{microtype}
\usepackage{subcaption}
\usepackage{times}
\usepackage{latexsym}
\usepackage{caption}
\usepackage{graphicx}

\usepackage{nccmath}
\usepackage{amssymb}
\usepackage{bbold}
\usepackage{comment}
\usepackage{pifont}% http://ctan.org/pkg/pifont
\aclfinalcopy

\newcommand{\TaBERT}{{\sc TaBert}}
\newcommand{\TAPAS}{{\sc TaPas}}
\newcommand{\RCI}{{\sc RCI}}

\newcommand{\cmark}{\ding{51}}%
\newcommand{\xmark}{\ding{55}}%

\title{\projName: An End-to-End, Transformer-Based System for Cell Level Table Retrieval and Table Question Answering}

\author{
{Feifei Pan$^{1}$}, {Mustafa Canim $^{2}$}, {Michael Glass $^{2}$}, {Alfio Gliozzo$^{2}$}, {Peter Fox$^{1}$}\\
{\tt panf2@rpi.edu}, {\tt mustafa@us.ibm.com}, \\{\tt mrglass@us.ibm.com}, {\tt gliozzo@us.ibm.com}\\{\tt pfox@cs.rpi.edu}
\\
$^{1}$ Rensselaer Polytechnic Institute \\
$^{2}$ IBM TJ Watson Research Center {\tt } \\
 \\
}

\date{}
\usepackage{xcolor}
\newcommand{\projName}{CLTR}
\newcommand{\bmNameGNQ}{E2E\_GNQ}
\newcommand{\bmNameWTQ}{E2E\_WTQ}

\newcommand{\eat}[1]{}
\begin{document}
\maketitle
%Accepted papers will be given one additional page of content so that reviewers’ comments can be taken into account.

\begin{abstract}
We present the first end-to-end, transformer-based table question answering (QA) system that takes natural language questions and massive table corpus as inputs to retrieve the most relevant tables and locate the correct table cells to answer the question \footnote{System page: https://github.com/IBM/row-column-intersection}. Our system, CLTR, extends the current state-of-the-art QA over tables model to build an end-to-end table QA architecture. This system has successfully tackled many real-world table QA problems with a simple, unified pipeline. Our proposed system can also generate a heatmap of candidate columns and rows over complex tables and allow users to quickly identify the correct cells to answer questions. In addition, we introduce two new open-domain benchmarks, E2E\_WTQ and E2E\_GNQ, consisting of 2,005 natural language questions over 76,242 tables. The benchmarks are designed to validate CLTR as well as accommodate future table retrieval and end-to-end table QA research and experiments. Our experiments demonstrate that our system is the current state-of-the-art model on the table retrieval task and produces promising results for end-to-end table QA.
\end{abstract}
\section{Introduction} \label{sec:intro}
\begin{figure*}
\centering 
\includegraphics[width=\linewidth]{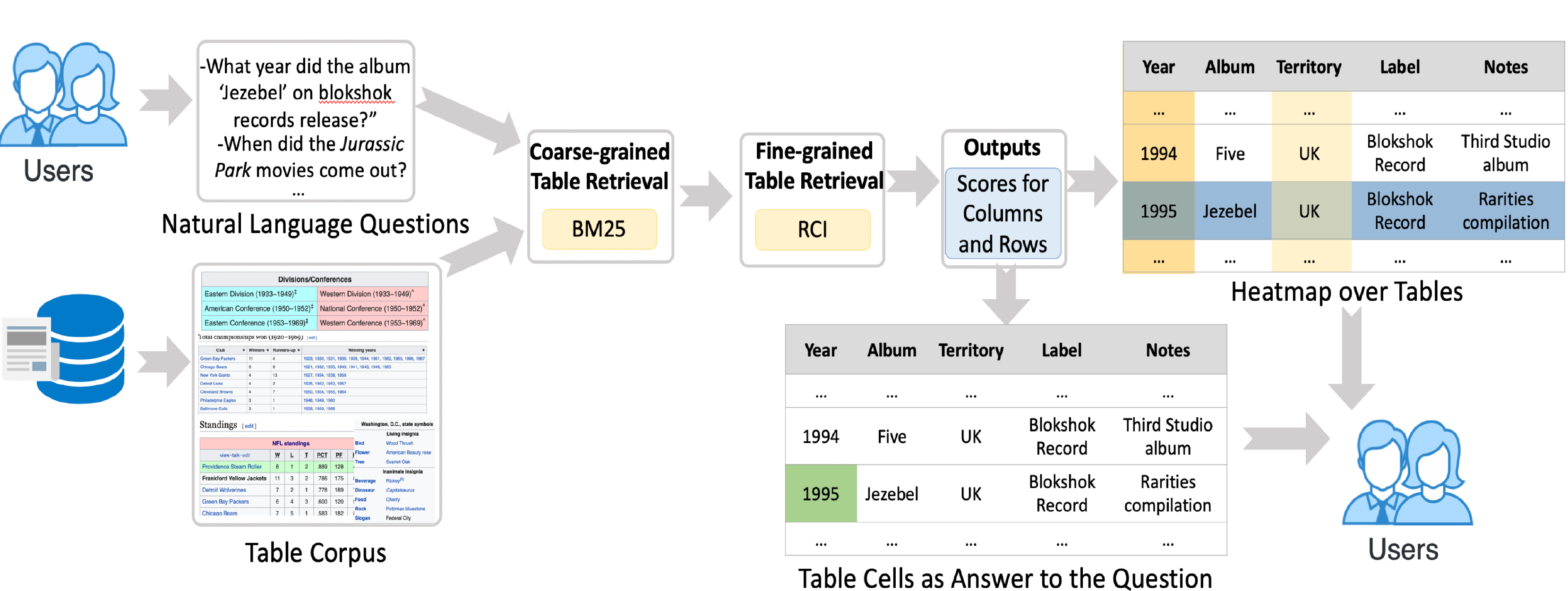}
\caption{The overview of the end-to-end table QA architecture of \projName.} 
\label{fig.overall}
\vspace{-3 mm}
\end{figure*}
Tables are widely used in digital documents across many domains, ranging from open-domain knowledge bases to domain-specific scientific journals, enterprise reports, to store structured information in tabular format. Many algorithms have been developed to retrieve tables based on given queries \cite{webtables, cafarella2009data, yan2017contentbased, Bhagavatula2013MethodsFE, shraga2020ad, chen2021open}. The majority of these solutions exploit traditional information retrieval (IR) techniques where tables are treated as documents without considering the tabular structure. However, these retrieval methods often result in an inferior quality due to a major limitation that most of these approaches highly rely on lexical matching between keyword queries and table contents. Recently, there is a growing demand to support natural language questions (NLQs) over tables and answer the NLQs directly, rather than simply retrieving top-$k$ relevant tables for keyword-based queries. \citet{sigir20} introduce the first NLQ-based table retrieval system, which leverages an advanced deep learning model. Although it is a practical approach to better understand the structure of NLQs and table content, it only focuses on table retrieval rather than answering NLQs. Lately, transformer-based pre-training approaches have been introduced in \TaBERT{} \cite{tabert}, \TAPAS{} \cite{tapas}, and the Row-Column Intersection model (\RCI{}) \cite{rci_inreview}. These algorithms are very powerful at answering questions on given tables; however, one cannot apply them over all tables in a corpus due to the computationally expensive nature of transformers. 
% Camera ready-
An end-to-end table QA system that accomplishes both tasks is in need as it has the following advantages over separated systems: (1) It reduces error accumulations caused by inconsistent, separated models; (2) It is easier to fine-tune, optimize, and perform error analysis and reasoning on an end-to-end system; and (3) It better accommodates user needs with a single, unified pipeline. Hence, we propose a table retrieval and QA over tables system in this paper, called \textbf{C}ell \textbf{L}evel \textbf{T}able \textbf{R}etrieval (\emph{\projName}). It first retrieves a pool of tables from a large table corpus with a coarse-grained but inexpensive IR method. It then applies a transformer-based QA over tables model to re-rank the table pool and finally finds the table cells as answers. To the best of our knowledge, this is the first end-to-end framework where a transformer-based, fine-grained QA model is used along with efficient coarse-grained IR methods to retrieve tables and answer questions over them. Our experiments demonstrate that \projName\ outperforms current state-of-the-art models on the table retrieval task while further helping customers find answers over returned tables. %While the IR-based component improves the scalability of the system, the powerful transformer model further improves the ranking quality of the retrieved results and identifies cell values over tables containing the answer of the questions.

To build such a Table QA system, an end-to-end benchmark is needed to evaluate alternative approaches. Current benchmarks, however, are not designed for such tasks, as they either focus on the retrieval task over multiple tables or QA task on a single table. To address the problems, we propose two new benchmarks: \bmNameWTQ\ and \bmNameGNQ. The details of these benchmarks and more discussions are provided in Section ~\ref{sec:data}. %These benchmarks are created by extending and manually annotating questions in \textit{WikiTableQuestions} dataset  \cite{pasupat2015compositional} and the \textit{GNQtables} dataset  \cite{sigir20}. As a result of this process we created two benchmarks that are rich enough to be used as part of a training process of an LTR task for the table retrieval process as well as the assessment of QA models over the given tables.

The specific contributions of this paper are summarized as follows:
\begin{itemize}
    \vspace{-1 mm}
    \item \textbf{A transformer-based end-to-end table QA system:} We build a novel end-to-end table QA pipeline by utilizing a transfer learning approach to retrieve tables from a massive table corpus and answer questions over them. The end system outperforms the state-of-the-art approaches on the table retrieval task.
    \vspace{-3 mm}
    \item \textbf{Creating heatmaps over complex tables:} To highlight all relevant table columns, rows, and cells, \projName\ generates heatmaps on tables. Following a pre-defined color code, the highlighted columns, rows, and cells are ranked according to their relevance to the questions. Using the heatmap, users can efficiently glance through complex tables and accurately locate the answers to the questions.
    \vspace{-3 mm}
    \item \textbf{Two new benchmarks for the end-to-end table QA evaluation:} We propose and release two new benchmarks, \bmNameWTQ\ and \bmNameGNQ, extending two existing benchmarks, \textit{WikiTableQuestions} and \textit{GNQtables}, respectively. The benchmarks can be used to evaluate systems for table retrieval and end-to-end table QA. 
\end{itemize} 

% In the following sections, we first describe the overall architecture of the system we propose in Section~\ref{sec:framework}. In Section ~\ref{sec:rci}, we discuss the RCI framework and our propose system, \projName. We validate our system in Section~\ref{sec:experiment} starting with introducing the end-to-end evaluation benchmarks and the experimental setup and metrics in Section~\ref{sec:data} and~\ref{sec:expSetup}, respectively. The experiment results are reported in Section~\ref{sec:expRes}, and finally, we review the previous work on existing table retrieval and QA technologies in Section~\ref{sec:relatedWork} conclude the paper in Section~\ref{sec:conclusion}.

\section{Overview} \label{sec:framework}
%TO-DO: Better define fine grained and coarse grained
\paragraph{The Architecture}
The architecture of our end-to-end table QA system, \projName, is illustrated in Figure~\ref{fig.overall}. This system aims to solve the end-to-end table QA task by generating a reasonable-sized subset of relevant tables from a massive table corpus, and employs the transformer-based approach to re-rank them based on their relevance to the user given NLQs, and finally answer the given NLQs with cells from these tables. 

\projName\ possess an abundant number of tables generated from documents of various knowledge sources to form a large table corpus. The system has two components: an inexpensive \textit{tf-idf} ~\cite{salton1986introduction} based coarse-grained table retrieval component and a fine-grained RCI-based table QA component. \projName\ first takes as input any user given NLQs and processes the questions and the table corpus with the inexpensive BM25 algorithm to generate a set of relevant tables, which is relatively large and contains noise (i.e., irrelevant tables). Here we use BM25 to efficiently narrow down the table candidates from a massive table corpus and highly reduce the execution time and computational cost for \projName. The output of this coarse-grained table retrieval component is later fed into the more expensive but accurate, transformer-based RCI to learn probability scores for table columns and rows, respectively. The scores produced by RCI indicate how likely the given question's final answer exists within a table column or row. With the probability scores, \projName\ re-ranks the tables and produces two outputs to the users: (1) a heatmap over top-ranked tables that highlights the most relevant columns and rows with a color code; (2) the table cells that contain the answers to the NLQs. 
\begin{figure*}
% \vspace{-3 mm}
\centering 
\includegraphics[width=\linewidth]{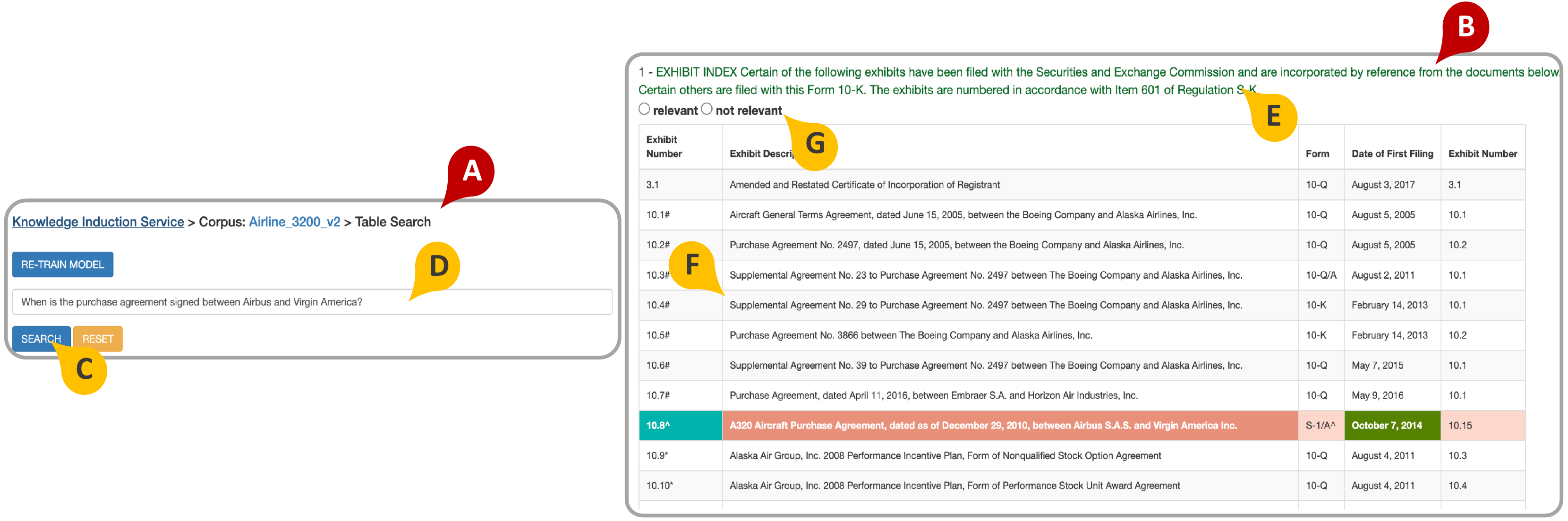}
\vspace{-3 mm}
\caption{The application of \projName\ on an aviation corpus} 
\label{fig.UI}
\end{figure*}

\paragraph{The applications}
Figure ~\ref{fig.UI} presents the user interface of an application of the \projName\ system. In this example, we apply the system to table QA over an aviation-related dataset, a domain-specific dataset on tables in aviation companies' annual reports. This user interface consists of two major sections, with \textit{Tag A} and \textit{Tag B} point to the user input and the system output sections, respectively. Under \textit{Tag A} and \textit{B}, the \projName\ pipeline is employed to support multiple functionalities. Users can input any NLQs, such as ``When is the purchase agreement signed between Airbus and Virgin America?'' in this example, into the text box at \textit{Tag D} and click the \textit{Search} button at \textit{Tag C} to query the pre-loaded table corpus. Users may select to reset the system for new queries or re-train a new model with a new corpus. In the system output sections, a list of tables similar to the table at \textit{Tag F} is generated and presented to users. For each table, the system output includes: (a) the surrounding text of the table from the original PDF (\textit{Tag E}); (b) the pre-processed table in a clean, human-readable format with a heatmap on it, indicating the most relevant rows, columns, and cells (\textit{Tag F}); (c) an annotation option, where the users can contribute to refining the system with feedback (\textit{Tag G}). In addition, the \projName\ architecture has been widely applied to datasets from many other domains, varying from finance to medical. The system is also validated with open-domain benchmarks, with more details discussed in Section \ref{sec:experiment}. 

\section{The RCI-based Table QA}\label{sec:rci}
Traditional approaches solve the table QA problem with two consecutive steps: retrieval of the most relevant tables for a given NLQ and locating the correct answers out of the cells with the help of a QA over tables model. These steps are usually studied separately. Our proposed system, \projName, unifies the two-step table QA with a single pipeline by leveraging the novel RCI model. RCI is the state-of-the-art approach for locating answers over tables \cite{rci_inreview}; however, it is not designed to retrieve tables out of large table corpus. In this section, we describe how we build an end-to-end table QA system combining the strength of inexpensive IR methods and the RCI model.

\subsection{The Row-Column Intersection Model}
% \footnote{Due to the anonymity period of the conference, the RCI work has not been referenced. Here we provide a brief description of the RCI model for completeness. This subsection will be updated with the correct reference in the camera-ready version of the paper.}
We first briefly introduce the Row-Column Intersection model (RCI), which supports the fine-grained table retrieval component of our system. The RCI model decomposes table QA into its two components: projection, corresponding to identifying columns, and selection, identifying rows.
Every row and column identification is a binary sequence-pair classification. 
The first sequence is the question and the second sequence is the row or column textual sequence representation.
We use the interaction model of RCI that concatenates the two sequences, with standard separator tokens, as the input to a transformer.  

The RCI interaction model uses the sequence representation which is later appended to the question with standard $[CLS]$ and $[SEP]$ tokens to delimit the two sequences. This sequence pair is fed into a transformer encoder, ALBERT~\cite{Lan2020ALBERT}. The final hidden state for the $[CLS]$ token is used in a linear layer followed by a softmax to classify if the column or row containing the answer or not. Each row and column is assigned with a probability of containing the answer. The \RCI{} model outputs the top-ranked cell as the intersection of the most probable row and the most probable column.
%Figure~\ref{fig.RCIinteraction} shows the interaction model of RCI.

\begin{figure}[thb]
  \centering
  \includegraphics[width=\linewidth]{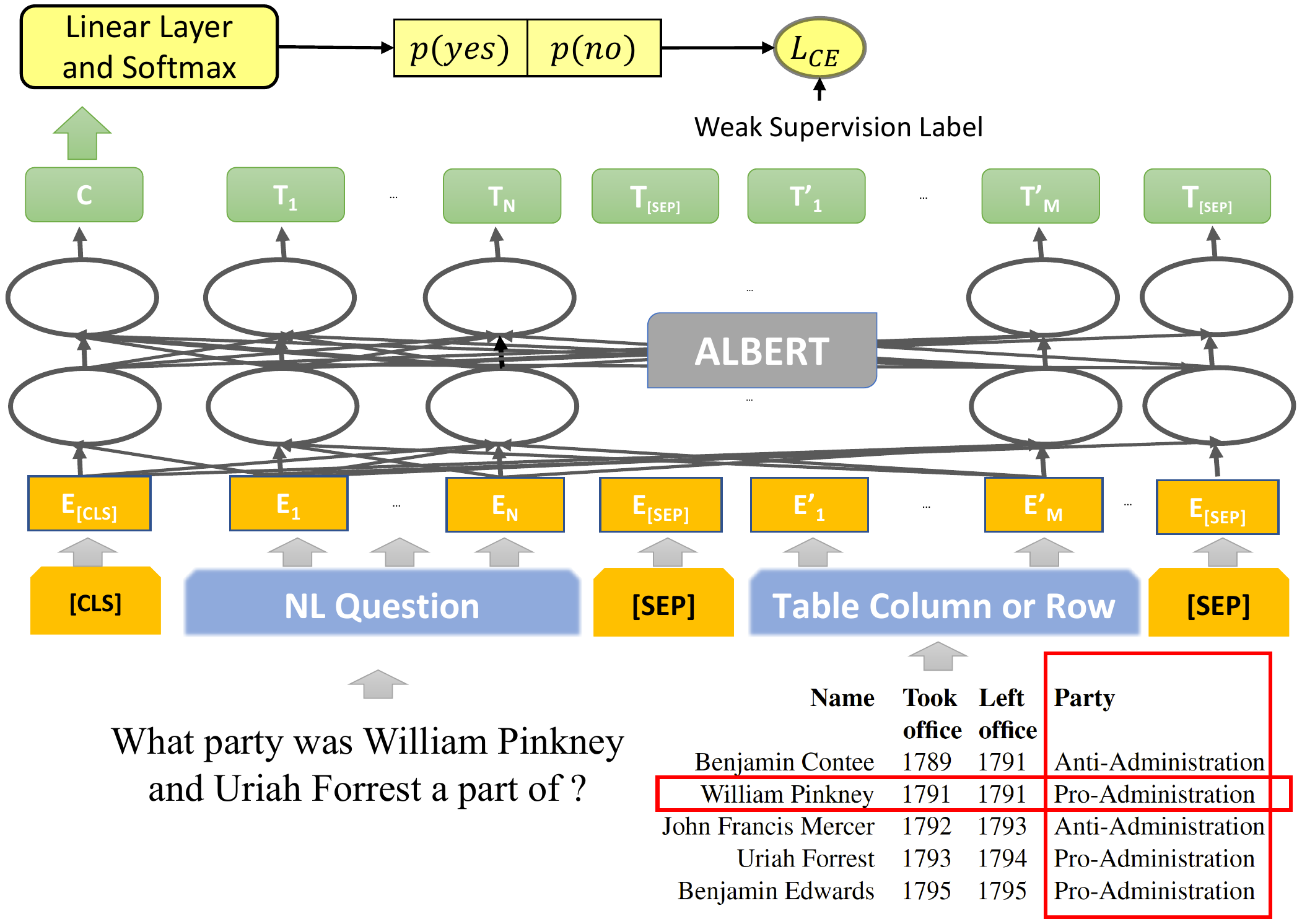}
  \caption{The RCI Table QA Model}
  \vspace{-1 mm}
  \label{fig.RCIinteraction}
  \vspace{-3 mm}
\end{figure}

Figure~\ref{fig.RCIinteraction} gives a sample question fed into the transformer architecture along with the column and row representation of a table.
%Here we formally describe how \RCI{} is used in the table retrieval and QA tasks.Let a table with $m$ rows and $n$ columns be defined as a header, $H = [h_1, h_2, ..., h_{n}]$ and cell values $V = [v_{i,j}], 1 \leq i \leq m, 1 \leq j \leq n$.  A table QA instance consists of a table, a question and a ground truth set of cell indices, $T \subseteq I \times J, I = {1,2,...,m}, J= {1,2,...,n}$. 

%To use the sequence-pair classification of a transformer on a table \RCI{} translates the structured row and column data to sequences. The row ($S^{r}_{i}$) and column ($S^{c}_{j}$) sequence representations are formatted as:

%\begin{equation}
%\begin{gathered}
%S^{r}_{i} = \bigoplus_{j=1}^{n} \zeta_h(h_j) \oplus %\zeta_v(v_{i,j}) \\
%S^{c}_{j} = \zeta_h(h_j) \oplus \bigoplus_{i=1}^{m} %\zeta_v(v_{i,j})
%\end{gathered}
%\end{equation}

%Where $\oplus$ indicates concatenation and the functions $\zeta_h$ and $\zeta_v$ delimit the header and cell value contents. For $\zeta_h$ we append a colon token (`:') to the header string, and for $\zeta_v$ we append a pipe token (`$|$') to the cell value string. The particular tokens used in the delimiting functions are not important. Any distinctive tokens can serve since the transformer will learn an appropriate embedding to represent their role as header and cell value delimiters.
\subsection{The End-to-End Table QA with RCI}\label{sec:end2end}
To tackle the table retrieval problem, we exploit an inexpensive IR method together with the state-of-the-art RCI model. Unlike the traditional methods treating tables as free text, a set of features, or multi-modal objects, \projName\ treats tables as a set of columns and rows and re-rank the tables based on cell-level RCI scores.

As we previously mentioned in Section ~\ref{sec:framework}, \projName\ first processes the question and table corpus with the inexpensive BM25 algorithm to generate a pool of highly relevant tables. Later, the RCI model is used to produce probability scores for every column and row for tables in the pool. Therefore, for every table $t$ with $n$ columns and $m$ rows in the table pool $T$, we have two set of scores, $P_{column} = \{p_{c_{1}}, p_{c_{2}}, p_{c_{3}},..., p_{c_{n}}\}$ for columns and $P_{row} = \{p_{r_{1}}, p_{r_{2}}, p_{r_{2}},..., p_{r_{m}}\}$ for rows. We calculate the overall probability score for each table by taking the maximum cell-level score, using $P_{t} = max(P_{col}) + max(P_{row})$. Our experiments prove the advantages of this method over the other algorithms (e.g., taking the averaged cell-level scores). 

\projName\ re-ranks the tables within the table pool $T$ using the maximum cell-level scores. Once the re-ranking is done, the top-$k$ tables out of $T$ are returned to the users. The correct cells on the top-$k$ tables are later identified by locating the intersection of the most relevant columns and rows discovered by the RCI model.

% \noindent \textbf{RCI:}   The Row-Column Intersection model (RCI) decomposes TableQA into its two components: projection, corresponding to identifying columns, and selection, identifying rows.
% Every row and column identification is a binary sequence-pair classification. 
% The first sequence is the question and the second sequence is the row or column textual sequence representation.
% We use the interaction model of RCI that concatenates the two sequences, with standard separator tokens, as the input to a transformer. 
\section{Experiments} \label{sec:experiment}
\subsection{Data}\label{sec:data} 
 \begin{table*}[ht]
 \footnotesize
        \centering
        \begin{tabular}{ccccccc}
          & \# of tables & \# of queries & Retrieval task & QA task &  Reference \\
        \hline\hline
WikiSQL & 24,241 &	80,654 &	\xmark & \cmark  & \cite{zhongSeq2SQL2017}\\
TabMCQ & 68 & 9,092 &	\xmark & \cmark & \cite{jauhar2015tabmcq}\\
WikiTableQuestions & 2,108 & 22,033 & \xmark & \cmark & \cite{pasupat2015compositional}\\ 
WikiTables & 1.6M & 100 & \cmark & \xmark & \cite{wikitable}\\ 
GNQtables &	74,224 &	789 & \cmark & \xmark & \cite{sigir20}\\ 
\hline
E2E\_WTQ &	2,108 &	1,216 & \cmark & \cmark\\ 
E2E\_GNQ &	74,224 & 789 & \cmark & \cmark\\ 
       \end{tabular}
       \vspace{-2 mm}
       \caption{Comparison of table QA and retrieval benchmarks}
     \label{tab:benchmarks}
     \vspace{-5 mm}
\end{table*}
\paragraph{Proposed Benchmarks: } Existing table retrieval and QA benchmarks focus on either answering NLQs on a single table or the retrieval of multiple tables for a keyword query. 
A comprehensive comparison of existing benchmarks with their limitations is listed in Table~\ref{tab:benchmarks}. WikiSQL \cite{zhongSeq2SQL2017} and WikiTableQuestions \cite{pasupat2015compositional} are widely used to evaluate table QA systems. More recently, they have been used by \TAPAS~\cite{tapas} and \TaBERT~\cite{tabert} where transformer-based models for QA over tables have been introduced. However, these benchmarks are not created to be used as part of an end-to-end table retrieval and QA pipeline. On the other hand, WikiTables was created based on the corpus introduced by \citet{wikitable} and used in many recent table retrieval studies \cite{Zhang2018AdHT, Deng2019Table2VecNW, www20, sigir20}. Despite its popularity, the WikiTables benchmark has two major limitations. First, the query set is fairly limited, containing only 100 keyword-based queries. Many recent studies use this small set of queries for a learning-to-rank (LTR) task with 5-fold cross-validation, potentially causing overfitting issues for the proposed table retrieval models. Second, the query set includes only keyword-based queries, which do not represent the NLQs customers are expected to ask to get answers over tables. To solve the aforementioned issues and create an end-to-end table QA benchmark with NLQs, we introduce two new benchmarks, \bmNameWTQ\ and \bmNameGNQ, inspired by \textit{WikiTableQuestions} and \textit{GNQtables}. 

The \textit{WikiTableQuestions} ~\cite{pasupat2015compositional} benchmark is originally designed for finding answer to questions from given tables. It consists of complex NLQs and tables extracted from Wikipedia. We filter the benchmark following ~\citet{rci_inreview} to generate a subset of 1,216 questions with 2,108 tables. 

The \textit{GNQtables} dataset, introduced in \citet{sigir20}, extends the Google Natural Questions (NQ) benchmark \cite{47761}. It contains 789 NLQs and a large table corpus of 74,224 tables. For each question, the ground truth only points to the most relevant table (with a binary grade 1 indicates \textit{relevant}), while all other tables in the table corpus are considered \textit{irrelevant} (grade 0). \textit{GNQtables} is the only table retrieval benchmark using NLQs, which makes it possible to adapt it to end-to-end table QA. To create the \bmNameGNQ, we manually annotate and enhance \textit{GNQtables} with additional ground truth data for each question: (1) the table cells containing the correct answers; (2) the index of the target columns; (3) the index of the target rows.

\paragraph{Experimental Data:} 
% camera ready
We experiment with \bmNameWTQ\ to test the portability of \projName, in which we fine-tune the RCI model with two other table QA benchmarks. We utilize an open-domain benchmark, WikiSQL~\cite{zhongSeq2SQL2017}, and a domain-specific benchmark, TabMCQ ~\cite{jauhar2015tabmcq}. The WikiSQL dataset has 80,654 questions on 24,241 Wikipedia tables, while the TabMCQ is a much smaller dataset, with only 68 hand-crafted tables and 9,092 multiple-choice questions. 

\subsection{Experimental Setup}\label{sec:expSetup}
%TODO qualitative analysis
\paragraph{Overall Setup:} 
We test our system under two experimental settings for table retrieval: (1) We test \projName\ without task-specific training on \bmNameWTQ\ and fine-tune the RCI model with WikiSQL and TabMCQ; (2) To fairly compare against the state-of-the-art, we follow the experimental setup in \citet{sigir20} and fine-tune \projName\ with \bmNameGNQ. We implement 5-fold cross-validation on \bmNameGNQ, where 80\% of data is used for fine-tuning and 20\% is used for validation. For both \bmNameGNQ\ and \bmNameWTQ, we use BM25 as our baseline model, which is widely used in industry-scale IR systems. We test the end-to-end table QA capability of \projName\ with our newly proposed benchmarks. Since we are the first publicly accessible end-to-end table QA system, we do not have a baseline to fairly compare to for our end-to-end table QA experiments.
% Camera ready-

We implement the coarse-grained table retrieval with the BM25 algorithm embedded in the ElasticSearch python API for all of our experiments. This API can be accessed at \textit{https://elasticsearch-py.readthedocs.io/en/master/}. Each table is indexed as a single text document with the embedded English analyzer. For each question, we generate a pool of 300 tables with the highest BM25 similarity scores. Following the current state-of-the-art model in ~\citet{sigir20}, we set $k1 = 1.2$ and $b = 0.7$. The tables in the pool are later processed with the RCI model. 

Our experiments employ the RCI model with ALBERT XXL version~\cite{Lan2020ALBERT}. The RCI model is fine-tuned for different benchmarks with the following configurations: (1) training batch size = 128; (2) Number of epochs = 2; (3) Learning rate = 2.5e-5; and (4) maximum sequence length = 512.

The model and data for the experiments with CLTR are available at \textit{https://github.com/IBM/row-column-intersection}.
\vspace{-1 mm}
\paragraph{Evaluation metrics:}
For table retrieval evaluation, we use the three metrics from previous work \cite{zhang2018ad, sigir20} for the top-$k$ retrieved tables, namely precision (P) with $k \in \{5, 10\}$, normalized discounted gain (NDCG) with $k \in \{5, 10, 20\}$, and the mean average precision (MAP). For the end-to-end table QA tasks, we evaluate our proposed model following \citet{rci_inreview} with two commonly used metrics in the IR community, accuracy at top 1 retrieved answer (Hit@1) and the mean reciprocal rank (MRR).

All experimental results are evaluated with the TREC standard evaluation tool \cite{10.5555/1121636}. The source code of the TREC evaluation tool can be found at \textit{https://trec.nist.gov/trec\_eval/}.
%\paragraph{End-to-End Table QA:} We test the the end-to-end table QA capability of \projName\ with our newly proposed benchmark. Due to the fact that there is no publicly accessible end-to-end table QA models, we cannot compare our results against other approaches. Therefore, we only report our results.

\subsection{Experimental Results}\label{sec:expRes}
We experimentally compare \projName\ against the BM25 baseline and the current state-of-the-art model on table retrieval in this section. Furthermore, we test \projName\ with our proposed benchmarks on the end-to-end table QA task. 
% \vspace{-1 mm}
\begin{table*}[h]
    \footnotesize
    \begin{subtable}[h]{0.9\textwidth}
        \centering
              \begin{tabular}{lllllll}
          & P@5 & P@10 & N@5 & N@10 & N@20 & MAP \\
        \hline\hline
        BM25 &  0.5938 & 0.6587 & 0.5228 & 0.5356 & 0.5359 & 0.4704\\
        \projName &  \textbf{0.7437} & \textbf{0.8735} & \textbf{0.6915} & \textbf{0.7119} & \textbf{0.7321} & \textbf{0.5971}\\
        \end{tabular}
        \caption{\bmNameWTQ}
        \label{tab:wtq}
    \end{subtable}
    \hfill
    \hfill
    \begin{subtable}[h]{0.9\textwidth}
    \footnotesize
        \centering
        \begin{tabular}{lllllll}
          & P@5 & P@10 & N@5 & N@10 & N@20 & MAP \\
        \hline\hline
        BM25 & 0.0413 & 0.0242 & 0.1650 & 0.1764 & 0.1852 & 0.1601 \\
        $MTR_{point}$ & 0.1460 & 0.0767 & 0.6227 & 0.6349 & 0.6359 &  0.5920 \\
        $MTR_{pair}$ & 0.1826$^*$ & 0.0990$^*$ & 0.6945$^*$ & 0.7198$^*$ & 0.7220$^*$ & 0.6328$^*$\\
        \projName &  \textbf{0.2203} & \textbf{0.1660} & \textbf{0.7235} & \textbf{0.7402} & \textbf{0.7458} & \textbf{0.7176}\\
       \end{tabular}
       \caption{\bmNameGNQ}
       \label{tab:gnq}
    \end{subtable}
    \vspace{-2 mm}
     \caption{A comparison of \projName\ and the baselines (* indicates the current state-of-the-art numbers).}
     \label{tab:tableRetrieval}
\vspace{-5 mm}
\end{table*}
\paragraph{Table Retrieval:} We present the experimental results for table retrieval without task specific training on \bmNameWTQ\ in Table~\ref{tab:wtq}. Since the MTR model \cite{sigir20} is not available to us and this dataset has never been used in any published table retrieval work, we only compare our results to the coarse-grained BM25 baseline. The results indicate our proposed model outperforms the BM25 baseline with average improvements of 29.12\%, 33.94\% and 26.93\% on precision, NDCG, and MAP, respectively. %Camera ready
The results on \bmNameWTQ\ also indicate that pre-trained \projName\ can be adapted to new datasets without task-specific training. 

The experimental results for \bmNameGNQ\ are shown in Table \ref{tab:gnq}, comparing against BM25 and the current state-of-the-art, the two MTR models, $MTR_{point}$ (with point-wise training) and $MTR_{pair}$ (with pair-wise training) in \citet{sigir20}. The comparison shows that our proposed model outperforms the current best $MTR_{pair}$ model on all metrics, with an average improvement of 28.73\% on precision, 3.43\% on NDCG, and 13.40\% on MAP. The experimental results indicates \projName\ is the new state-of-the-art system for table retrieval. Moreover, \projName\ can further locate cell values to answer NLQs after table retrieval.
%%%%%% pvalue=0.9236700746764042
\begin{table}[h]%{0.45\textwidth}
\footnotesize
\vspace{-2 mm}
    \centering
    \begin{tabular}{lll}
    & MRR& Hit@1\\
    \hline\hline
    \bmNameWTQ\ & 0.5503 & 0.4675\\
    \bmNameGNQ\ & 0.4067 & 0.2699\\
    \end{tabular}
    \vspace{-2 mm}
    \caption{Model evaluation for end-to-end table QA}
    \label{tab:e2e}
\vspace{-6 mm}
\end{table}

\paragraph{End-to-End Table QA:} 
To further validate \projName, we implement the end-to-end Table QA evaluation with \bmNameWTQ\ and \bmNameGNQ. The only existing end-to-end table QA model, \citet{sun2016table}, and its dataset are not publicly available. Therefore, we do not have any baseline models to compare to. Our experimental results are reported in Table ~\ref{tab:e2e}. As the first attempt for an end-to-end table QA system with transformer-based architecture on complex table benchmarks, we show that our approach is able to achieve promising and consistent performance. Our results indicate \projName\ performs better for the first benchmark, \bmNameWTQ, where the table corpus mainly contains well-structured tables. On the other hand, we expect the results for \bmNameGNQ\ to be worse due to the amount of poorly formatted tables in the table corpus. 

\paragraph{Qualitative Analysis: } %Camera ready
The experiments indicate \projName\ outperforms all baselines, as well as the current state-of-the-art models on table retrieval. It also produces promising results for the end-to-end table QA task. We further demonstrate the high-portability of \projName\ with pre-trained models using unseen benchmarks. 

The system performance is much better for \bmNameWTQ\ based on the experimental results. After a thorough investigation, we notice that the original \textit{GNQtables} contains a large amount of noisy tables which do not have tabular structures. A considerable amount of tables in \textit{GNQtables} are Wikipedia \textit{InfoBoxes}, which may have multiple column/row headers and are difficult to process by machines accurately. Although table quality is crucial for table QA models, \projName\ proves its advantageous by producing state-of-the-art results with noisy table corpus. Furthermore, the example shown in Figure~\ref{fig.UI} demonstrates the effectiveness of \projName\ when applied to real-world data.

\section{Related Work} \label{sec:relatedWork}
\paragraph{Table Retrieval}
A majority of the table retrieval methods proposed in the literature treat tables as individual documents without taking the tabular structure into consideration ~\cite{pyreddy1997tintin, wangandhu, liu2007tableseer, webtables, cafarella2009data}.
More recent approaches utilize features generated from queries, tables, or query-table pairs. For example, \citet{zhang2018ad} introduces an ad-hoc table retrieval method, retrieving tables with features such as \#query\_term, \#columns, \#null\_values, etc. Similar work includes \citet{yan2017contentbased}, \citet{Bhagavatula2013MethodsFE}, and \citet{shraga2020ad}. 
The current state-of-the-art model is introduced in \citet{sigir20}, where tables are treated as multi-modal objects and retrieved with a neural ranking model. We compare \projName\ with this approach in Section \ref{sec:experiment}.
\vspace{-1 mm}
\paragraph{Table QA Models}
Early table QA systems typically convert natural language questions into SQL format to answer questions over tables \cite{yu-etal-2018-spider, guo2020content, lin2019grammarbased, xu2018sqlnet}. 
In \citet{JimenezRuizHEC20}, the authors promote the idea of matching tabular data to knowledge graphs and create the Semantic Web Challenge on Tabular Data to Knowledge Graph Matching (SemTab), which provide a new solution for table understanding and QA related tasks.
Recently, \TAPAS~\cite{tapas} and \TaBERT~\cite{tabert} introduce the transformer-based approaches for this task.
The \RCI~\cite{rci_inreview} model is the state-of-the-art model for QA over tables. It utilizes a transfer learning based framework to independently classify the most relevant columns and rows for a given question and further identify the most relevant cells as the intersections of top-ranked columns and rows. 
\vspace{-1 mm}
\paragraph{End-to-End Table QA Models} 
To the best of our knowledge, the table cell search framework published in \citet{sun2016table} is the only existing end-to-end Table QA system. This work leverages the semantic relations between table cells and uses relational chains to connect queries to table cells. However, the proposed model only works for well-formatted questions containing at least one highly relevant entity to link tables to the questions. In addition, the model and the data are not publicly available for comparison. 
\section{Conclusion} \label{sec:conclusion}
This paper proposes an end-to-end solution for table retrieval and finding answers for NLQs over tables. To the best of our knowledge, this is the first system built where a transformer-based QA model is used for locating answers over tables while improving the ranking of tables out of a table pool formed by inexpensive IR methods. To evaluate the efficacy of this system, we introduce two benchmarks, namely \bmNameWTQ\ and \bmNameGNQ.

The experimental results indicates that the proposed system, \projName, outperforms the baselines and the current state-of-the-art model on the table retrieval task. Furthermore, \projName\ produces promising results on the end-to-end table QA task. In real-world applications, \projName\ can be applied to create a heatmap over tables to assist users in quickly identifying the correct cells on tables.

\bibliographystyle{acl_natbib}
\bibliography{anthology,acl2021}

%\appendix

\end{document}